%% file: main.tex
\documentclass{article}
\usepackage{arxiv}

\usepackage{hyperref}       

\usepackage{amsfonts}       

\usepackage{graphicx}
\usepackage{caption}
\usepackage{subcaption}
\usepackage{amsmath, amssymb}

\title{Topology Optimization using Neural Networks with Conditioning Field Initialization for Improved Efficiency}

\date{} 					

\author{Hongrui Chen \qquad Aditya Joglekar \qquad \textbf{Levent Burak Kara}\thanks{Address all correspondences to lkara@cmu.edu} \\ Department of Mechanical Engineering\\Carnegie Mellon University\\ Pittsburgh, PA, 15213, USA}

\begin{document}

\maketitle    

\begin{abstract}
{
We propose conditioning field initialization for neural network based topology optimization. In this work, we focus on (1) improving upon existing neural network based topology optimization, (2) demonstrating that by using a prior initial field on the unoptimized domain, the efficiency of neural network based topology optimization can be further improved. Our approach consists of a topology neural network that is trained on a case by case basis to represent the geometry for a single topology optimization problem. It takes in domain coordinates as input to represent the density at each coordinate where the topology is represented by a continuous density field. The displacement is solved through a finite element solver. We employ the strain energy field calculated on the initial design domain as an additional conditioning field input to the neural network throughout the optimization. The addition of the strain energy field input improves the convergence speed compared to standalone neural network based topology optimization.  }
\end{abstract}

\input{s1}

\input{s2}

\input{s3}

\input{s4}

\input{s5}

\input{s6}

\bibliographystyle{unsrt}

\input{main.bbl}
\end{document}

%% file: s1.tex
\section{INTRODUCTION}
\input{fig_flowchart}

There has been a recent increase in machine learning driven topology optimization approaches, particularly using neural networks for performing topology optimization. Both data-driven and online training based approaches have been explored. Data-driven approaches require large training database generation and a long training time. They perform instant optimal topology generation during inference time. Online training approaches use the neural network to represent the density field of a single to a small subset of designs for better parameterization. 
The online training approaches require similar or more time compared to conventional topology optimization approaches like SIMP (Solid Isotropic Material with Penalisation) \cite{bendsoe1989optimal,zhou1991coc}. We find that the results of the online training approaches, particularly the convergence speed, can be improved through insights derived from the mechanical aspects of the problem.

Machine learning driven topology optimization approaches offer the advantage of being easily able to accommodate additional insights in the form of pre-computed fields. The usage of these fields has been explored in data-driven approaches such as TopologyGAN  \cite{nie2021topologygan}, which use physical fields such as von Mises stress and strain energy density for achieving better results. However, there has been no work incorporating these physical fields in the online training topology optimization setting. In this work, we further improve upon TOuNN (Topology Optimization using Neural Networks), an online training approach proposed by Chandrasekhar and Suresh \cite{Chandrasekhar2021}, by adding a strain energy field in addition to the domain coordinates as a conditioning input to the neural network. We show that this improves the convergence speed and can give a better compliance. With the additional strain energy field as a conditioning input, the neural network not only learns a mapping function between the domain coordinates to the density field output but also between the strain energy field to the density field output. Ideally, if the conditioning field is the same as the converged topology, then the neural network only needs to learn a constant function which is the identity function. 
However, the converged topology is not known at the beginning of the optimization. Thus, the strain energy field is used as a good alternative since it can be computed through a single function call of Finite Element Analysis (FEA) prior to the online training of the neural network. We verify the performance increase obtained with this additional conditioning input across parametric experiments with varying boundary conditions and volume fractions.

The code for running the experiments in this paper can be found at: https://github.com/HongRayChen/Hybrid-TopOpt

%% file: fig_flowchart.tex
\begin{figure*}
\centering

\includegraphics[width=\textwidth]{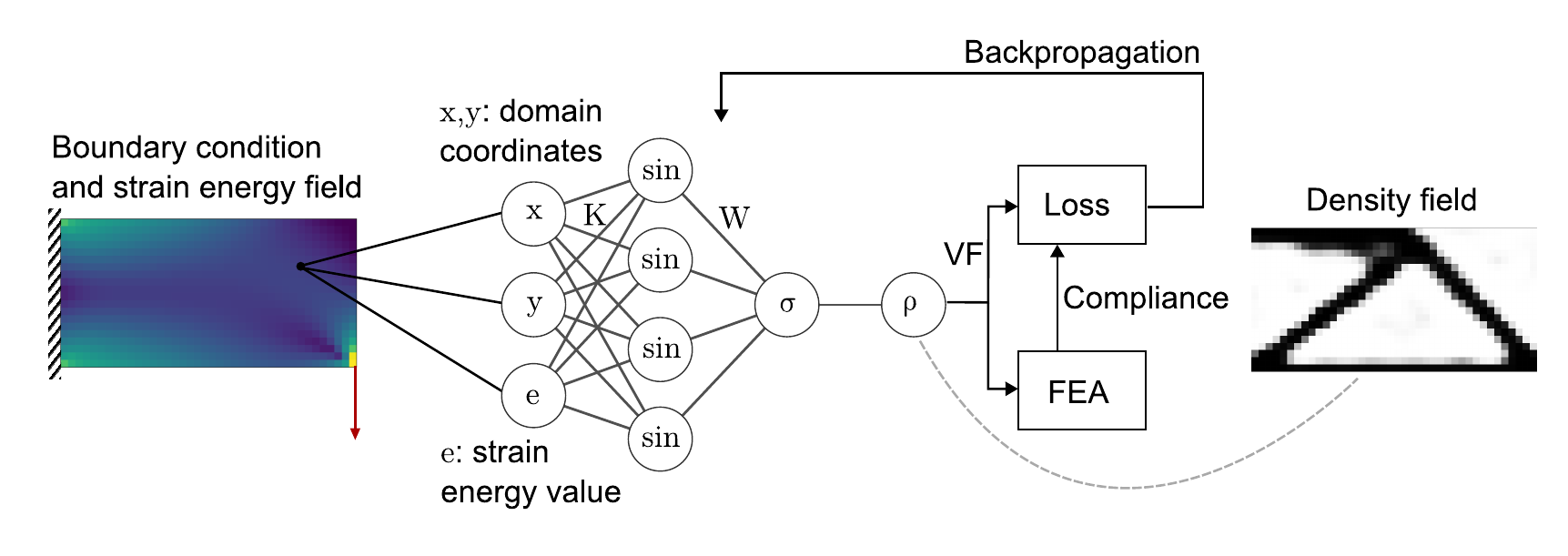}

\caption{The strain energy field is calculated at the beginning of the optimization based on the boundary condition. The strain energy conditioning field is fixed throughout the training. Domain coordinates and the strain energy value at each coordinate point is used as the input to the neural network. The neural network outputs density $\rho$ at each coordinate point. By sampling coordinate point across the design domain, we obtain the density field. From the density field, we calculate the current volume fraction and the compliance from a FEA solver. The compliance and volume fraction is then formulated as a loss function which is used in back propagation of the training process until convergence. }

\label{fig:flowchart}
\end{figure*}

%% file: s2.tex
\section{RELATED WORK}
\textit{Conventional topology optimization}:
Bends{\o}e and Kikuchi \cite{bens0e1988generating} introduced the homogenization approach for topology optiimization. The SIMP method \cite{bendsoe1989optimal,zhou1991coc}  considers the relative material density in each element of the Finite Element (FE) mesh as design variables, allowing for a simpler interpretation and optimised designs with more clearly defined features. Other common approaches to topology optimization include the level-set method \cite{allaire2002level,wangm2003guo} and evolutionary algorithms \cite{xie1997basic}.

All these methods use an iterative process to create a complex mapping from problem characteristics (supports, loads and objective function) to an optimised structure, where each iteration has an expensive FEA calculation involved. A more accurate and detailed solution can be obtained with greater number of elements in the FE mesh, however this increases the computational cost.  Therefore, current developments within the field are strongly motivated by the desire to either limit the number of iterations needed to obtain an optimised structure or the computational cost of completing an iteration \cite{woldseth2022use}. Recent advances in deep learning, particularly for image analysis tasks, have showed potential for removing the expensive FEA iterations required until the convergence of the topology in the conventional topology optimization approaches. Hence, various topology optimization approaches that utilize neural networks have been proposed. Woldseth et al. \cite{woldseth2022use} provide an extensive overview on this topic.

\textit{Data-driven topology optimization}: 
 We refer to data-driven topology optimization methods as those that aim to learn a neural network model from a database of topology optimization results for instant prediction of the optimal topology. Many methods rely on Convolutional Neural Networks (CNN) for their capabilities to learn from a large set of image data. Banga et al. \cite{banga20183d} used a 3D encoder-decoder CNN to generate 3D topology results and show that interpolating the final output using the 3D CNN from the initial iterations obtained from the ‘TopOpt’ \cite{aage2015topology} solver, offers a 40$\%$ reduction in time over the conventional approach of using the solver alone. Yu et al. \cite{yu2019deep} use a conditional generative adversarial network (cGAN) in addition to CNN based encoder-decoder network. However, the results indicate there sometimes there may be disconnections present in the predicted topology which may drastically affect the compliance values. Nakamura and Suzuki \cite{nakamura2020deep} improve on the results with their direct design network and with a larger dataset, however, disconnections are still observed in some solutions. Behzadi and Ilieş \cite{behzadi2021real} used deep transfer learning with CNN. Zheng et al. \cite{zheng2021generating} used U-net CNN for 3D topology synthesis. Nie at al. \cite{nie2021topologygan} used various physical fields computed on the original, unoptimized material domain, as inputs to the generator of a cGAN and achieved a 3 times reduction in mean square error as compared to a baseline cGAN. Maz{\'e} and Ahmed \cite{mazediffusion} show that diffusion models can outperform GANs for this task. They use regressor and classifier guidance to ensure that the generated structures are manufacturable and mechanical compliance has been minimized.
 
 All these data-driven approaches aim to reduce optimal topology prediction time but face difficulties in generalization. Though over the years there have been improvements on the generalization capability, suitable training dataset generation is not trivial, especially for the 3D domain, and satisfactory and reliable results have not been achieved yet for direct use in real-world problems.

\textit{Online training topology optimization}:
We refer to online training topology optimization methods as those which do not use any prior data, rather train a neural network in an self-supervised manner for learning the optimal density distribution/topology. Chandrasekhar and Suresh \cite{Chandrasekhar2021} explored a online approach where the density field is parameterized using a neural network. 
Fourier projection based neural network for length scale control \cite{Chandrasekhar2021Fourier} and application for multi-material topology optimization \cite{Chandrasekhar2021MM} has also been explored . Deng and To \cite{deng2020topology} propose topology optimization with Deep Representation Learning, with a similar concept of re-parametrization, and demonstrate the effectiveness of proposed method on minimum compliance and stress-constrained problems. Deng and To \cite{deng2021parametric} also propose a neural network based method for level-set topology optimization, where the implicit function of level-set is described by a fully connected deep neural network. Zehnder et al. \cite{zehnder2021ntopo} effectively leverage neural representations in the context of mesh-free topology optimization and use multilayer perceptrons to parameterize both density and displacement fields. It enables self-supervised learning of continuous solution spaces for topology optimization problems. Mai et al. \cite{mai2023physics} develop a similar approach for optimum design of truss structures. Hoyer et al. \cite{hoyer2019neural} use CNNs for density parametrization and directly enforce the constraints in each iteration, reducing the loss function to compliance only. They observe that the CNN solutions are qualitatively different from the baselines and often involve simpler and more effective structures. Zhang at al. \cite{zhang2021tonr} adopt a similar strategy and show solutions for different optimization problems including stress-constrained problems and compliant mechanism design. 

Generalization is not an issue with all these online training topology optimization methods. However, the computational time and cost is similar to traditional topology optimization approaches. An advantage offered is that the density representation is independent of the FE mesh and because of the analytical density-field representation, sharper structural boundaries can be obtained \cite{Chandrasekhar2021}. We show that by adding an initial condition field as an extra input, we can improve the convergence speed and get better results.

%% file: s3.tex
\input{fig_SE_norm}
\input{fig_SE_vf}

\section{PROPOSED METHOD}
In our proposed method, the density distribution of the geometry is directly represented by the topology neural network. The strain energy field and the compliance used for backpropagation is calculated from an FE solver. The program is implemented in Python and backpropagation of the loss function into each module is handled by the machine learning package TensorFlow \cite{Abadi2016}.


%
\subsection{Neural network}
The topology network $T(\textbf{X})$  (Figure \ref{fig:flowchart}), learns a density field in a different manner as compared to typical topology optimization which represents the density field as a finite element mesh. The topology neural network takes in domain coordinates $x,y$, as well as the strain energy value $e$ at coordinate $x,y$. The strain energy value gets concatenated with the domain coordinates to form the input to the topology network, $\textbf{X} = [x,y,e]$. The domain coordinates are normalized between $-0.5$ to $0.5$ for the longest edge. It outputs the density value $\rho$ at each coordinate point. The domain coordinates represent the center of each element in the design domain. During topology optimization, a batch of domain coordinates that correspond to the mesh grid and the corresponding strain energy field is fed into the topology network. The output is then sent to the Finite Element Analysis (FEA) solver. The solver outputs the compliance which is combined with the volume fraction violation as a loss. The loss is then backpropagated to learn the weights of the topology network. 

For the topology network design, we employed a simple architecture that resembles the function expression of $f(x) = \textbf{w}sin(\textbf{k}x+\textbf{b})$. Similar neural network architectures have been used to control the length scale of geometry in topology optimization\cite{Chandrasekhar2021Fourier}. The conditioned domain coordinates are multiplied with a kernel $\textbf{K}$. The kernel $\textbf{K}$ regulates the frequency of the sine function. We add a constant value of 1 to break the sine function's rotation symmetry around the origin. We use a Sigmoid function to guarantee the output is between 0 and 1. The topology network can be formulated as follows:
\begin{equation}
T(\textbf{X}) = \sigma(\textbf{W}\sin(\textbf{K}\textbf{X} +  1))
\end{equation}

\noindent where:

$\textbf{X}$: Domain coordinate input, $\textbf{X}=(x,y,e)$ 

$\sigma$: Sigmoid activation function

$\textbf{K}$: Trainable frequency kernels, initialized in $[-25,25]$ 

$\textbf{W}$: Trainable weights, initialized to 0

\noindent We can upsample the 3D coordinate input or only sample specific regions of the density field to manipulate the resolution of the discretized visualization. Due to the strain energy conditioning field computed from the finite element mesh grid, interpolation needs to be used to calculate the intermediate values when upsampling the domain coordinates. 

\subsection{Strain energy conditioning field}
The strain energy conditioning field is used to augment the domain coordinate input. We calculate the conditioning field from the initial homogeneous density domain. In topology optimization, for a 2D problem with n elements of four nodes each, the strain energy field $\textbf{E}$ can be calculated as follows:


\begin{equation}
\textbf{E} = \sum{(\textbf{U}_e \times \textbf{S}_e) \circ \textbf{U}_e}
\end{equation}

\noindent where:

$\textbf{U}_e$: the displacement matrix, $n \times 8$

$\textbf{S}_e$: the element stiffness matrix, $8 \times 8$

The summation is along the axis containing the values for each element.

In most topology optimization implementations, the compliance is then calculated by summation of the above strain energy for all elements. 

The strain energy field can vary greatly in range depending on the problem domain size, boundary condition, and geometry constraints. Therefore, normalization needs to be done to regulate the value range of the strain energy field. Otherwise, the range of the strain energy field will deviate from the normalized range of the domain coordinates. Furthermore, a simple normalization will not suffice as the high max value of the strain energy field reduces the amplitude of other relevant features and patterns (Figure \ref{fig:SE_norm} (a)). We explore gamma and logarithmic filtering to normalize the strain energy field. For the gamma filtering, we clip the strain energy field by using the 99th percentile, $P_{99}$. After clipping, more details of the field $\textbf{E}_{c}$ can be seen (Figure \ref{fig:SE_norm} (b)). We also further adjust the feature of the strain energy field by using gamma correction. The gamma value is set to be the complement of the target volume fraction $V^*$ for the optimization ($\gamma = 1 - V^*$). The effect of the gamma correction based on the volume fraction is illustrated in Figure \ref{fig:se_vf}. As the volume fraction increases, the edge feature in the strain energy field is more and more pronounced. Finally, after the gamma correction step, the strain energy field is normalized between 0 and 0.4 to obtain the processed field $\textbf{E}_p$. The processing step on the gamma filtering of strain energy field can be summarized in the following equation:

\begin{equation}
    \textbf{E}_{c} = min(\textbf{E} ,P_{99})
\end{equation}

\begin{equation}
    \textbf{E}_{\gamma} = 0.4 \Bigl\{\frac{\textbf{E}_{c} - min(\textbf{E}_{c})}{max(\textbf{E}_{c}) - min(\textbf{E}_{c})}\Bigr\} ^ {\gamma} 
\end{equation}

For the logarithmic filtering, we do not clip the value, instead, the log filter is directly applied to the strain energy field and then normalized between 0 and 0.4. We determine this range empirically to give the best results. 

\begin{equation}
    \textbf{E}_{\log} = 0.4\frac{\log\textbf{E} - min(\log\textbf{E})}{max(\log\textbf{E}) - min(\log\textbf{E})} 
\end{equation}

\input{fig_conv_beam}

\subsection{Online topology optimization with neural network }
During optimization, the topology network outputs the density value at the center for each element. These density values are then sent to the finite element solver to calculate compliance based on the SIMP interpolation. 

The finite element solver is treated as a black box within the neural network. It takes in the density of each element and outputs the compliance and the sensitivity for each element with respect to the compliance. Variables that are being optimized are the weights $\textbf{W}$ and kernels $\textbf{K}$ of the neural network. Adam\cite{kingma2014adam} is used to train the neural network. The constrained optimization problem needs to be transferred into unconstrained minimization problem for neural network. We adopt the loss function formulated by Chandrasekhar and Suresh \cite{Chandrasekhar2021} of compliance minimization and volume fraction constraint. The combined loss function is 
\begin{equation}
L = \frac{c}{c_0}+\alpha(\frac{\bar{\rho}}{V^*}-1)^2
\end{equation}

In the optimization, the target volume fraction $V^*$ is an equality constraint and $\bar{\rho}$ is the volume fraction of the current design. When $\alpha$ increased to infinity, the equality constraint is satisfied. We assign a maximum value of 100 for $\alpha$ with initial value of 1 and gradually increase $\alpha$ every iteration. $c$ is the current compliance and $c_0$ is the initial compliance calculated on the design domain with the uniform volume fraction $V^*$.

%% file: fig_SE_norm.tex
\begin{figure*}
\centering

\begin{subfigure}[t]{0.2\textwidth}
\centering
\includegraphics[width=\textwidth]{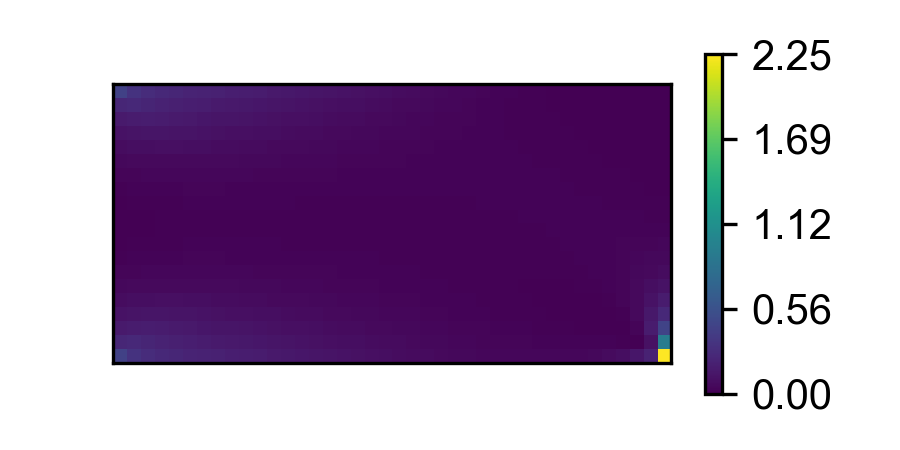}
\caption{Raw strain energy field output}
\end{subfigure}
\qquad
\begin{subfigure}[t]{0.2\textwidth}
\centering
\includegraphics[width=\textwidth]{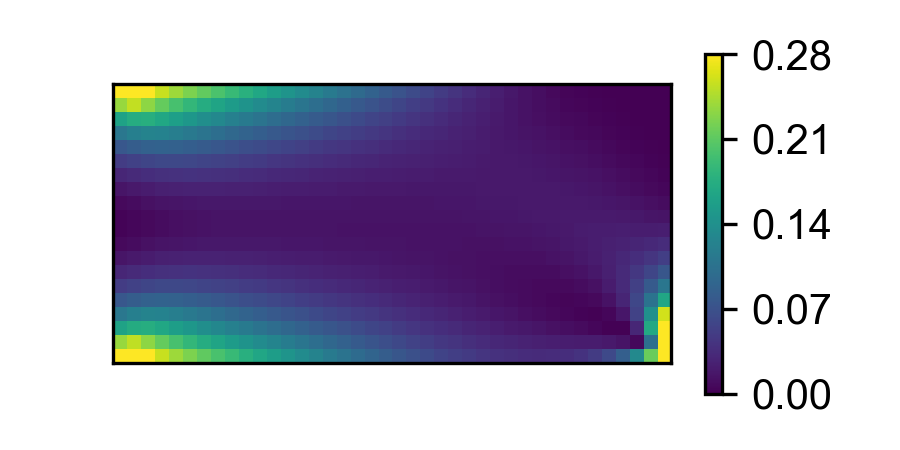}
\caption{clipping value above 99 percentile}
\end{subfigure}
\qquad
\begin{subfigure}[t]{0.2\textwidth}
\centering
\includegraphics[width=\textwidth]{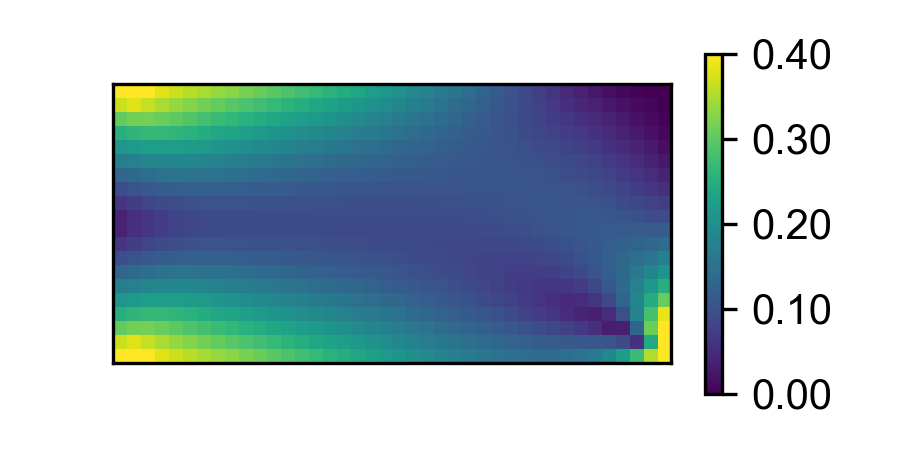}
\caption{conditioning field with gamma filter}
\end{subfigure}
\qquad
\begin{subfigure}[t]{0.2\textwidth}
\centering
\includegraphics[width=\textwidth]{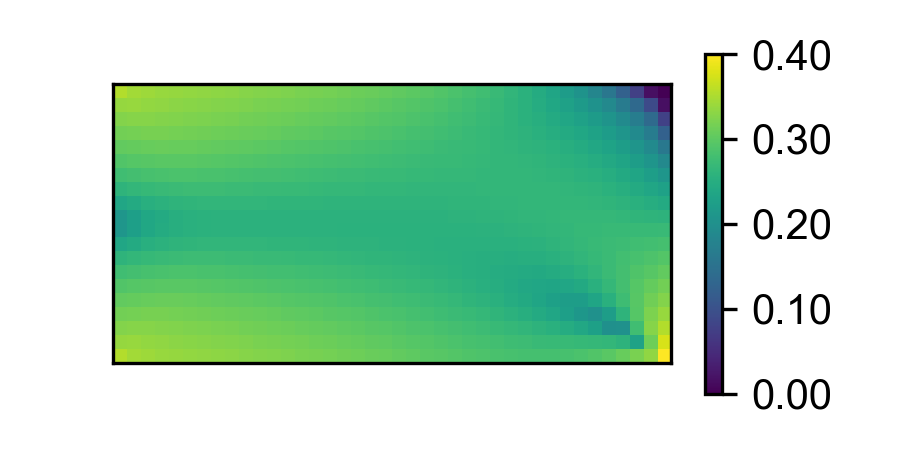}
\caption{conditioning field with log filter}
\end{subfigure}

\centering
\caption{Two method is evaluated in terms of processing of the strain energy conditioning field. We used a gamma filter in (a-c) and a log filter in (d)}

\label{fig:SE_norm}
\end{figure*}

%% file: fig_SE_vf.tex
\begin{figure*}
\centering

\includegraphics[width=\textwidth]{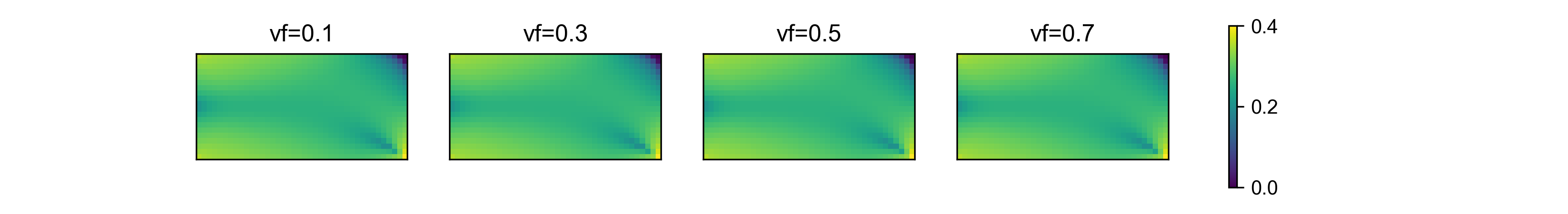}

\caption{For the gamma filtering of the conditioning field, we adjust the gamma based on the volume fraction target of the optimization}

\label{fig:se_vf}
\end{figure*}

%% file: fig_conv_beam.tex
\begin{figure*}
\centering
\begin{subfigure}[t]{0.25\textwidth}

\includegraphics[width=\textwidth]{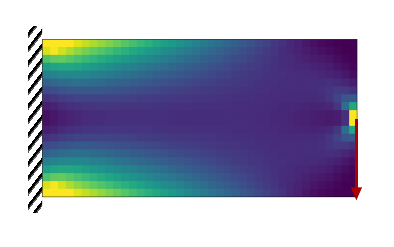}
\caption{Beam boundary condition}
\end{subfigure}
\qquad
\begin{subfigure}[t]{0.65\textwidth}

\includegraphics[width=\textwidth]{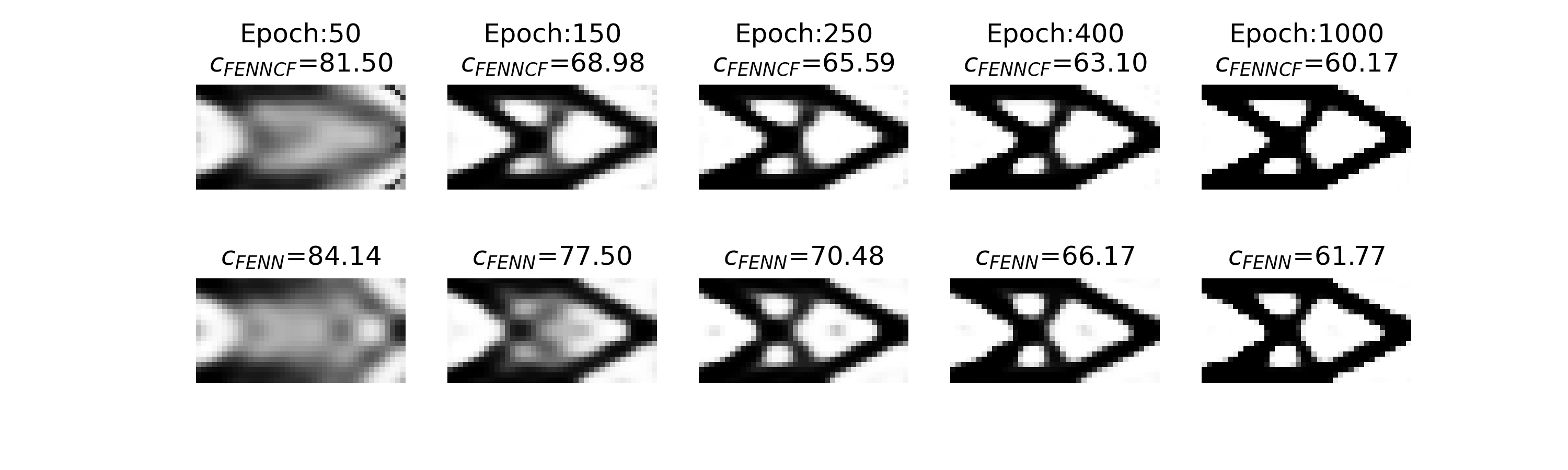}
\caption{Density field snapshot at 50, 150, 250, 400, and 1000 epoch}
\end{subfigure}

\begin{subfigure}[t]{0.9\textwidth}
\includegraphics[width=\textwidth]{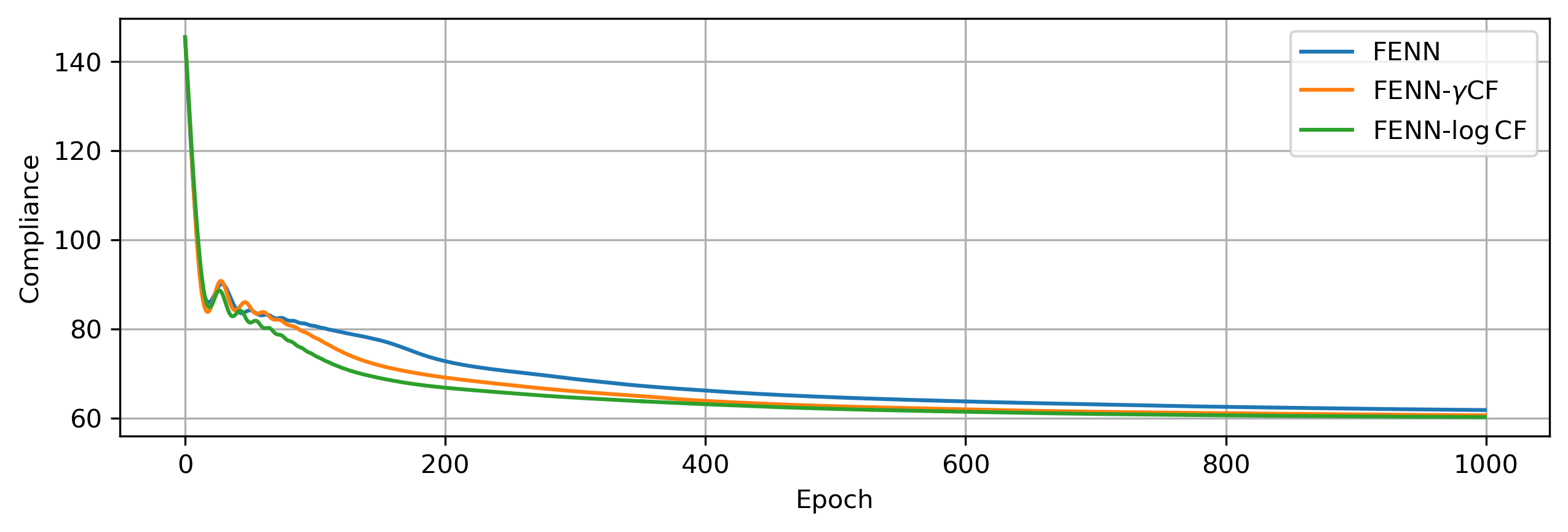}
\caption{Convergence history}
\end{subfigure}

\centering
\caption{Comparing the convergence history for a beam example for with and without strain energy conditioning field. The result presented is using the gamma filtering. For FENN-logCF took 22.5s while FENN took 22.1s. }

\label{fig:conv_beam}
\end{figure*}

%% file: s4.tex
\section{RESULTS AND DISCUSSIONS}
The possible combinations of boundary conditions, problem size, and configurations is enormous. It is impossible for us to cover all. To demonstrate the effectiveness of our proposed approach, we explore both a beam problem and a parametric study in 2D. In the beam problem, we showcase the convergence of the network's output and the convergence history. In the parametric study, problems across different boundary conditions and volume fractions are explored. We report the compliance value where subscript FENN represents Finite Element (FE) compliance solver with Neural Network (NN) as topology representation, and FENNCF as neural topology optimization with strain energy Conditioning Field (CF). For these two experiments, the problem size is 40$\times$20 pixels. All experiments are run on a PC with i7-12700K as processor, 32 GB of RAM, and Nvidia RTX3080 GPU.

\subsection{Beam example}
Our first experiment is the beam example. The left side of the domain is fixed and a downward point load is on the center-right side. The boundary condition illustration and the strain energy conditioning field are shown in Figure \ref{fig:conv_beam}. The target volume fraction is 0.3. We run the online topology optimization for a total of 1000 epochs. 

The convergence history plot is illustrated in Figure \ref{fig:conv_beam} (c). We observe that by epoch 50, with the strain energy conditioning field, the network's compliance takes over the lead and maintains lower compliance all the way to the end of the training epochs. We also show the density field snapshot and the corresponding compliance during training in Figure \ref{fig:conv_beam} (b). Analyzing the geometry of the neural network with the conditioning field, we observe that there is a subtle difference compared to without the conditioning field. The neural network with conditioning field shares greater similarities to the strain energy field where the top and bottom edges are shorter. We can also observe that most of the geometry convergence happens between 0 and 400 epochs. Between 400 to 1000 epochs, the geometry remained relatively unchanged. The only change being a darker tone of red, showing the density values get pushed closer towards 1. In both of the examples, the final volume fraction is within 1$\%$ error of the given target volume fraction. Therefore, we do not include the volume fraction convergence plot. 

\input{fig_c_compare_xPhys}
\input{fig_c_compare}

\subsection{Parametric study}

We set up a parametric study to analyze the effectiveness of the gamma and log filter of the conditioning field. The boundary condition setup is illustrated in Figure \ref{fig:c_compare_xPhys} (a). The bottom right loading point is varied across the region highlighted in green which accounts for 50 load conditions. We also vary the target volume fraction between 0.2 to 0.5 with an increment of 0.1. In total, this sums up to 200 total combinations. In the previous beam example, we observe that geometries do not change significantly after 400 epochs, therefore we limit the total epochs for the parametric study to 400 epochs. 

The parametric study result is summarized in Figure \ref{fig:c_compare}. In Figure \ref{fig:c_compare} (a), we sort with respect to the compliance of topology optimization without conditioning field and show the compliance from both methods. We observe that the overall conditioning field converged at lower compliance. The improvement of the conditioning field is more significant when the compliance is higher. The higher compliance occurs when the volume fraction is low. To visualize the convergence speed increase, Figure \ref{fig:c_compare} shows the percentage improvement with the conditioning field. The percentage improvement is calculated by identifying the epoch at which the conditioning field reaches a lower compliance compared to the final compliance of the optimization without the conditioning field. The average performance increase with gamma filter is $37.6\%$ and with log filter is $44.7\%$. With both filters, the performance increase is more pronounced with lower volume fraction examples. The log filter has a better overall performance increase across all solutions compared to gamma filter.

We compare our result against the result of "88-lines" by Andreassen et. al. \cite{andreassen2011efficient} with a filtering radius of 1.5 to accommodate the problem size. We observe that when the compliance is low, FENN performed slightly better than SIMP. This is also consistent with the result reported by Chandrasekhar and Suresh \cite{Chandrasekhar2021}. For problems with relatively higher compliance, we observe that FENN with conditioning field can in some cases converge to a lower compliance than "88-lines". We note that in general, the Matlab code \cite{andreassen2011efficient} takes around 0.2 to 1.5s to run whereas FENN and FENN with either conditioning field takes around 10s. However, a definite time comparison is difficult to establish as "88-lines" runs on Matlab whereas FENN runs on Python. In "88-lines" the optimizer is optimality criteria whereas FENN rely on Adam with a learning rate of 0.002. 

We also observe that within the 200 examples with gamma filter, there are four cases where the conditioning field does not improve convergence speed. When plotting out example results in Figure \ref{fig:c_compare_xPhys}, the examples with the load on the right bottom edge have lower performance increase with the conditioning field. On the other hand the examples with the load close to the center have a greater performance increase and a bigger gap in compliance. Our hypothesis is that the conditioning field approach performs best when the topology is complex.  The complexity in geometry can occur based on the volume fraction constraint or the configuration of the boundary conditions. As the volume fraction decrease, thinner members are required which increase complexity of the structure. Whereas the geometries in Figure \ref{fig:c_compare_xPhys} (b) showed that for the same volume fraction, the length scale of the part is also dependent on the boundary condition.

\input{fig_4_2Dcase}
\input{fig_case1_ss}
\input{fig_3D_CB_03}

\subsection{Additional examples}
In Figure \ref{fig:2D_case}, we demonstrate the improvements resulting from the conditioning field on 4 complex boundary conditions in 2D. Cases 2, 3 and 4 in Figure \ref{fig:2D_case} have obstacle regions (passive elements). Furthermore, in Figure \ref{fig:case1_ss}, we analyze the impact of increasing the problem resolution (i.e. the FE mesh size) for the boundary conditions of case 1 in Figure \ref{fig:2D_case}, and observe similar improvements. We also show the improvements seen for a 3D problem in Figure \ref{fig:CB_3D_03}.

%% file: fig_c_compare_xPhys.tex
\begin{figure*}

\centering

\begin{subfigure}[t]{0.25\textwidth}

\includegraphics[width=\textwidth]{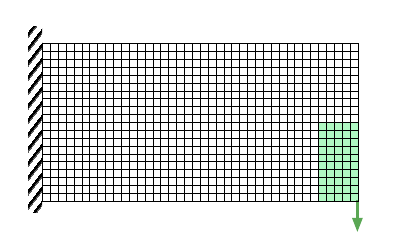}
\caption{Parameteric study boundary conditions}
\end{subfigure}
\qquad
\begin{subfigure}[t]{0.65\textwidth}
\includegraphics[width=\textwidth]{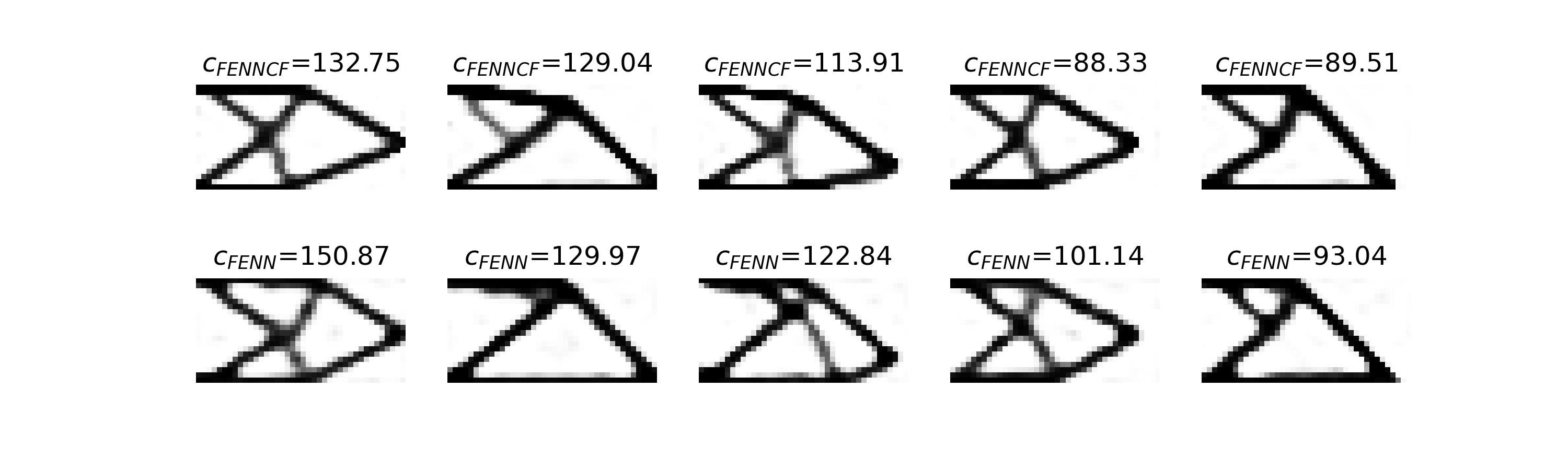}
\caption{Sample topology optimization with 0.3 volume fraction}
\end{subfigure}

\caption{Boundary conditions and some sample topology optimization results with 0.3 volume fraction within the parametric study examples}

\label{fig:c_compare_xPhys}
\end{figure*}

%% file: fig_c_compare.tex
\begin{figure*}
%
\begin{subfigure}[t]{0.45\textwidth}
\centering
\includegraphics[width=\textwidth]{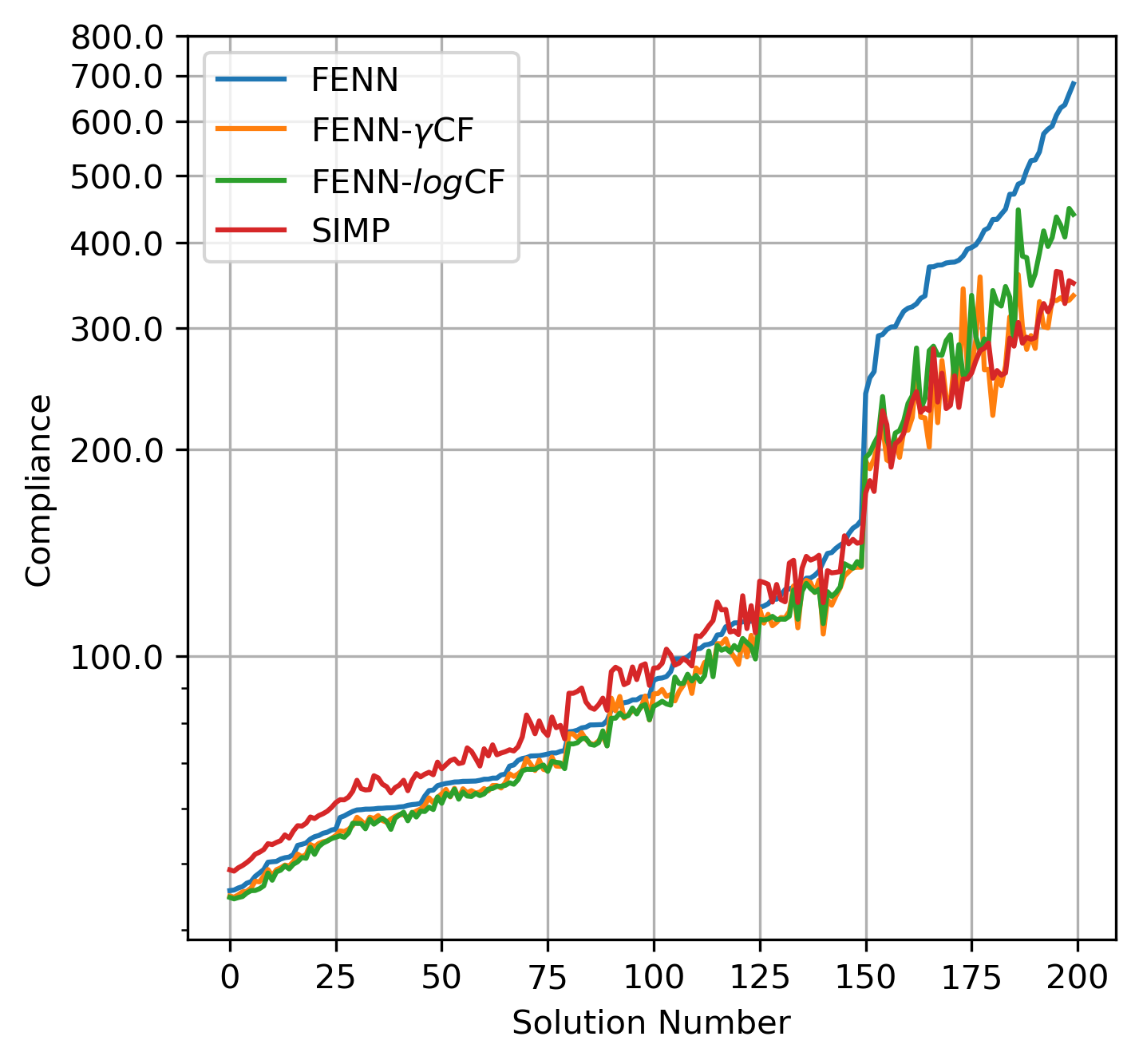}
\caption{Compliance comparison of with and without strain energy field}
\end{subfigure}
\qquad
\begin{subfigure}[t]{0.45\textwidth}
\centering
\includegraphics[width=\textwidth]{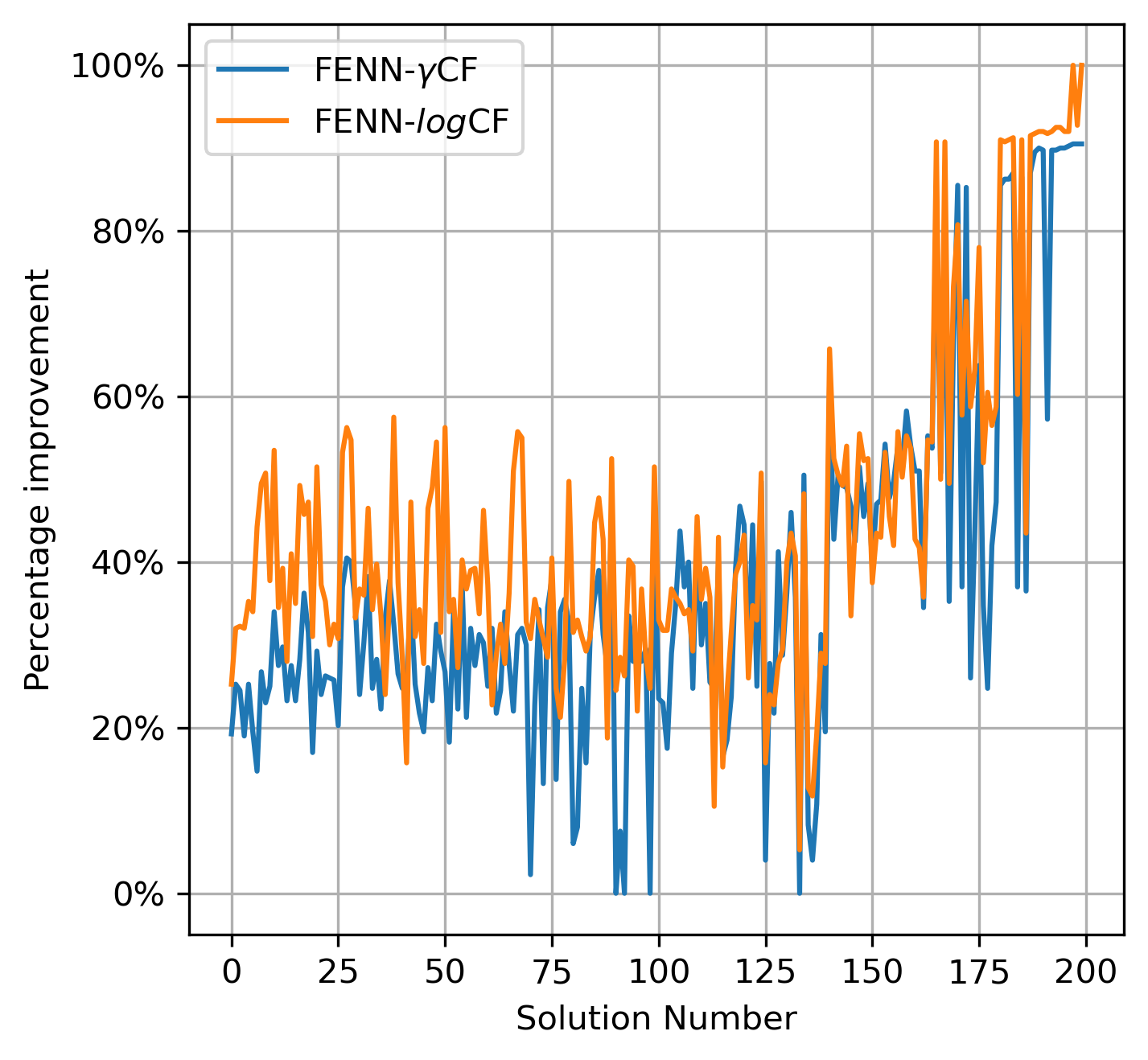}
\caption{Percentage convergence speed improvement}
\end{subfigure}

\centering
\caption{Comparing the final compliance and the speed of convergence for parametric study examples for with gamma and log filter and without the conditioning field. We also run the same problem configuration with "88-lines" by Andreassen et al. \cite{andreassen2011efficient} denoted by the legend "SIMP" in the figure.  }

\label{fig:c_compare}
\end{figure*}

%% file: fig_4_2Dcase.tex
\begin{figure*}[htbp]

Case 1
\hfill
\centering
\begin{subfigure}[C]{0.23\textwidth}
\includegraphics[width=\textwidth]{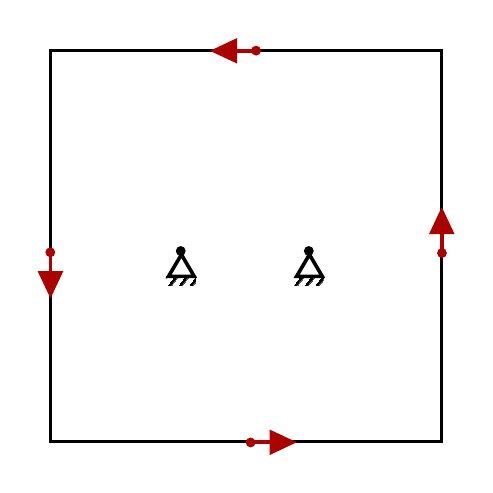}
\end{subfigure}
\qquad
\begin{subfigure}[C]{0.65\textwidth}
\includegraphics[width=\textwidth]{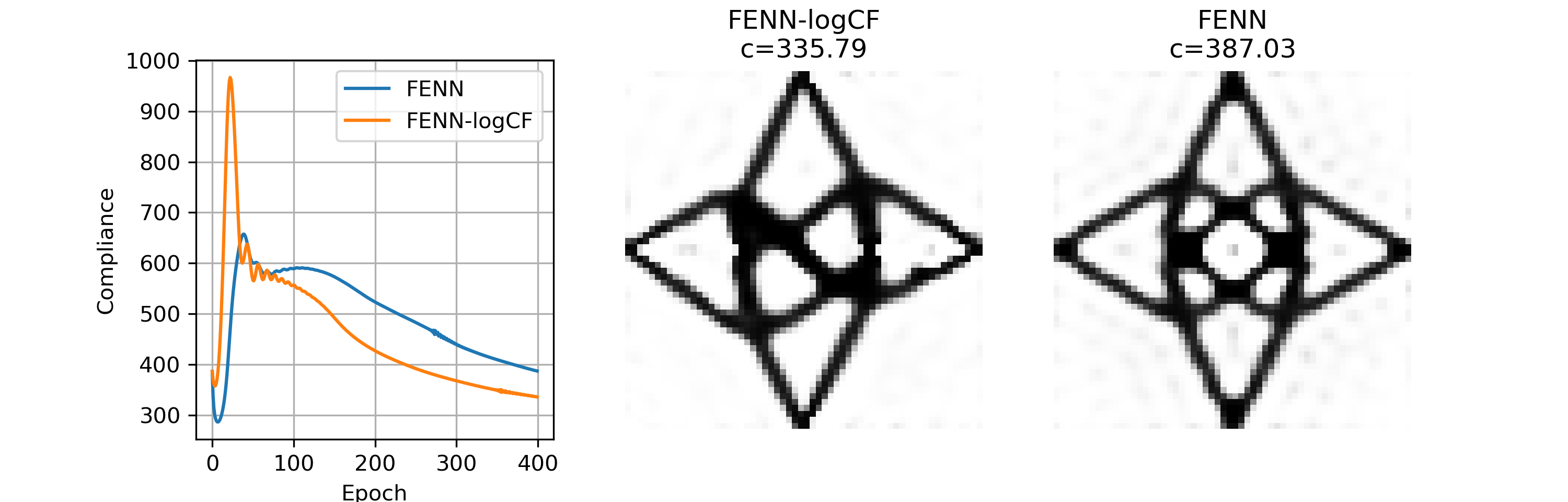}
\end{subfigure}

Case 2
\hfill
\centering
\begin{subfigure}[C]{0.23\textwidth}
\includegraphics[width=\textwidth]{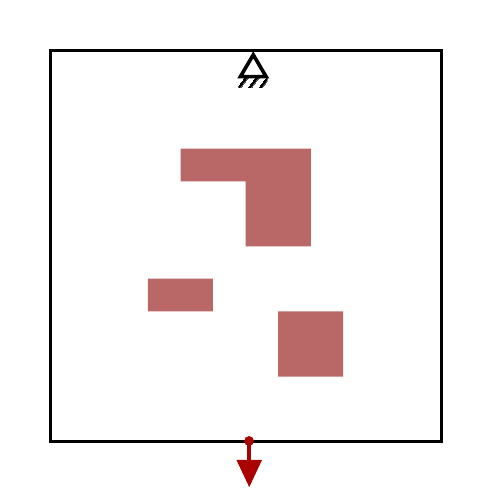}
\end{subfigure}
\qquad
\begin{subfigure}[C]{0.65\textwidth}
\includegraphics[width=\textwidth]{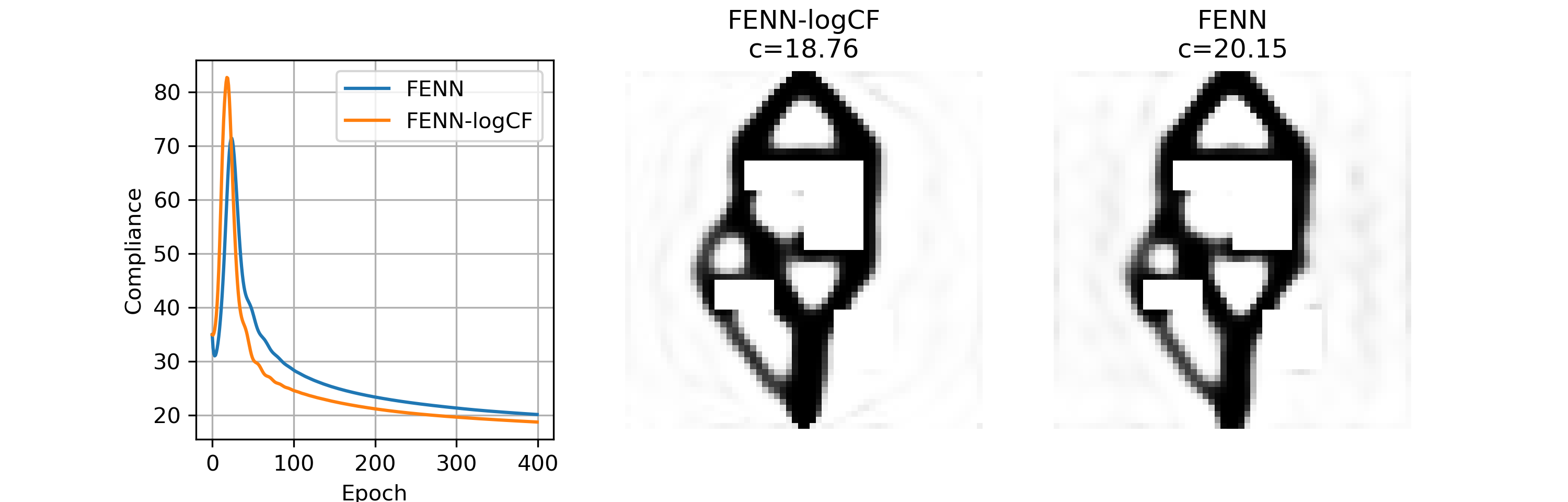}
\end{subfigure}

Case 3
\hfill
\centering
\begin{subfigure}[C]{0.23\textwidth}
\includegraphics[width=\textwidth]{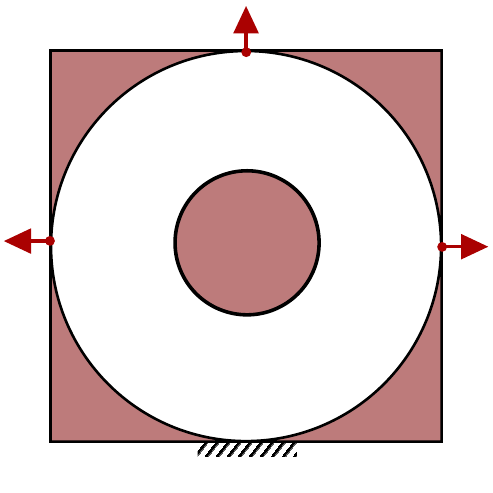}
\end{subfigure}
\qquad
\begin{subfigure}[C]{0.65\textwidth}
\includegraphics[width=\textwidth]{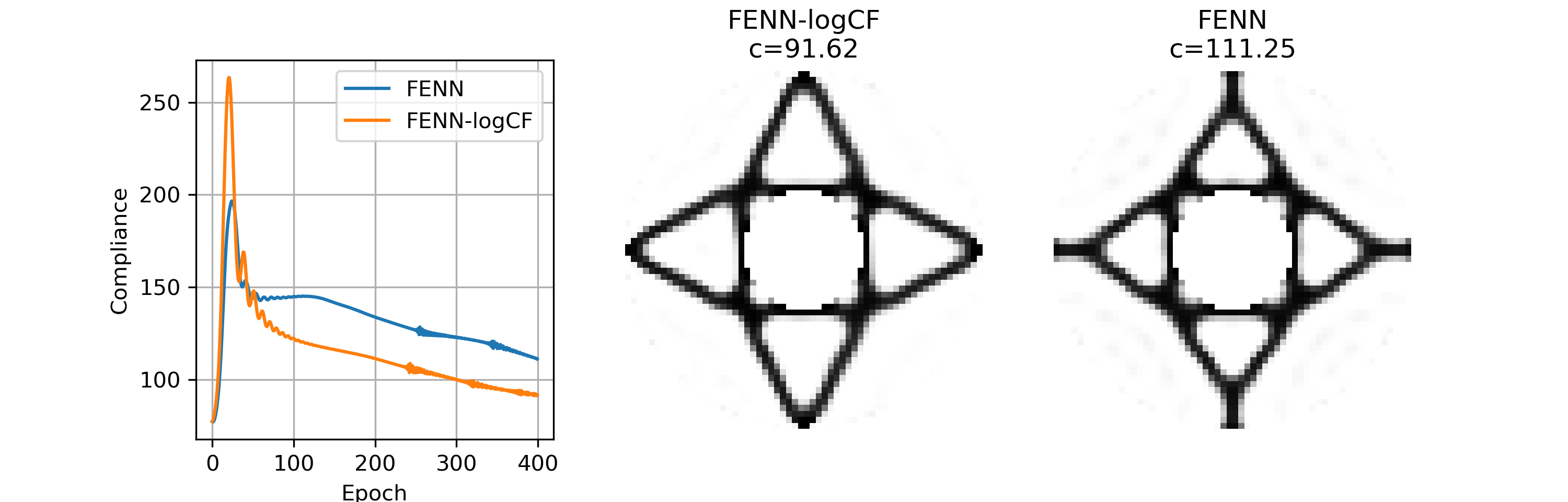}
\end{subfigure}

Case 4
\hfill
\centering
\begin{subfigure}[C]{0.23\textwidth}
\includegraphics[width=\textwidth]{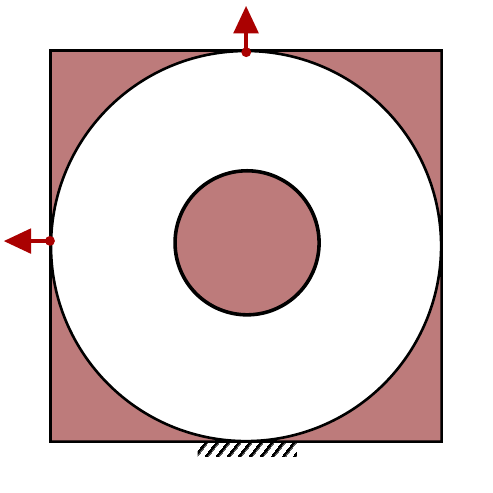}
\end{subfigure}
\qquad
\begin{subfigure}[C]{0.65\textwidth}
\includegraphics[width=\textwidth]{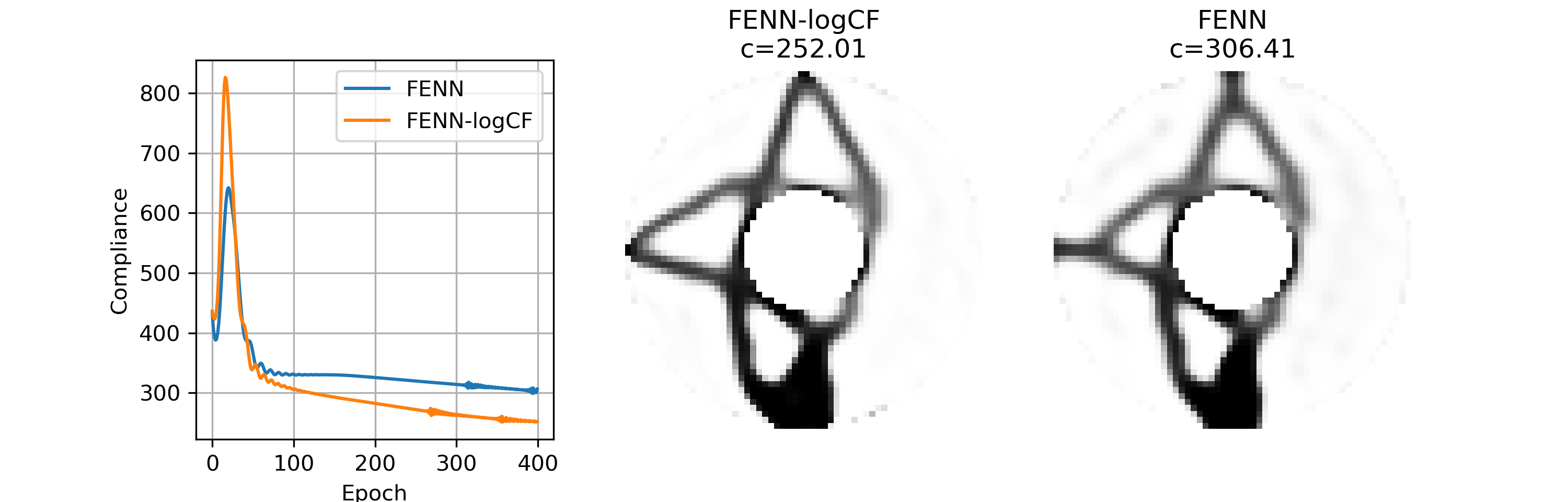}
\end{subfigure}

\caption{Four additional test cases across varying boundary conditions and passive elements, all using 0.2 target volume fraction. Each example is 60$\times$60 in resolution and takes around 30 seconds to run with no significant difference between with and without conditioning field. Log filtered conditioning field demonstrates good convergence speed increase.}

\label{fig:2D_case}
\end{figure*}

%% file: fig_case1_ss.tex
\begin{figure*}[htbp]

Case 1 with $120\times120$ resolution
\hfill
\centering
\begin{subfigure}[C]{0.65\textwidth}
\includegraphics[width=\textwidth]{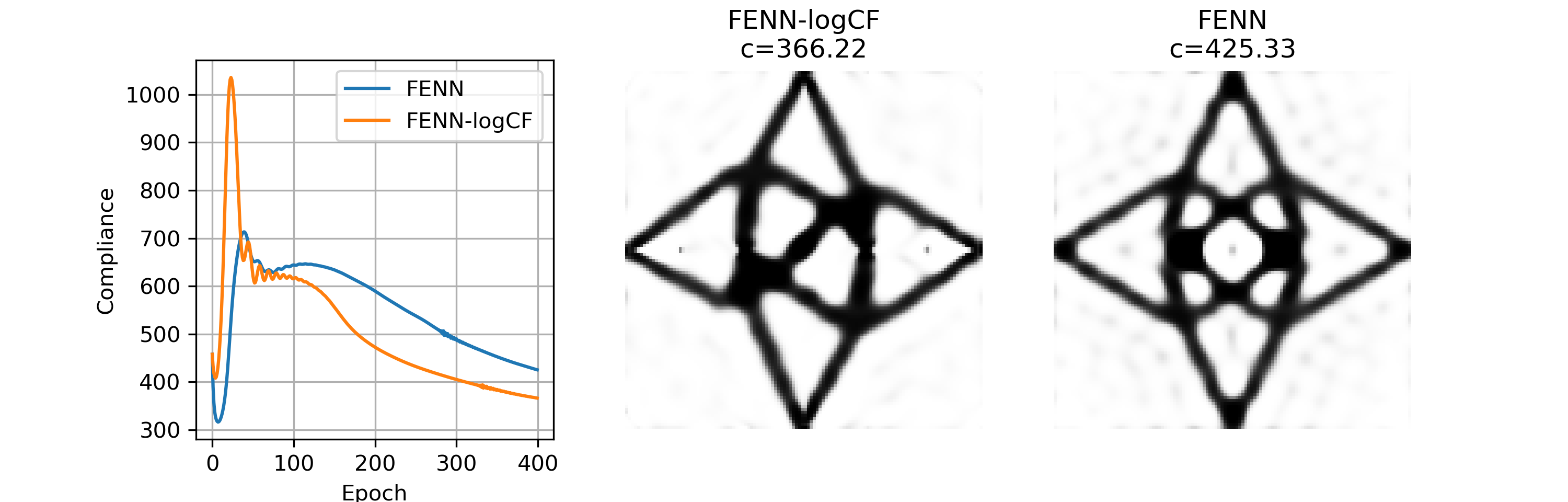}
\end{subfigure}

Case 1 with $180\times180$ resolution
\hfill
\centering
\begin{subfigure}[C]{0.65\textwidth}
\includegraphics[width=\textwidth]{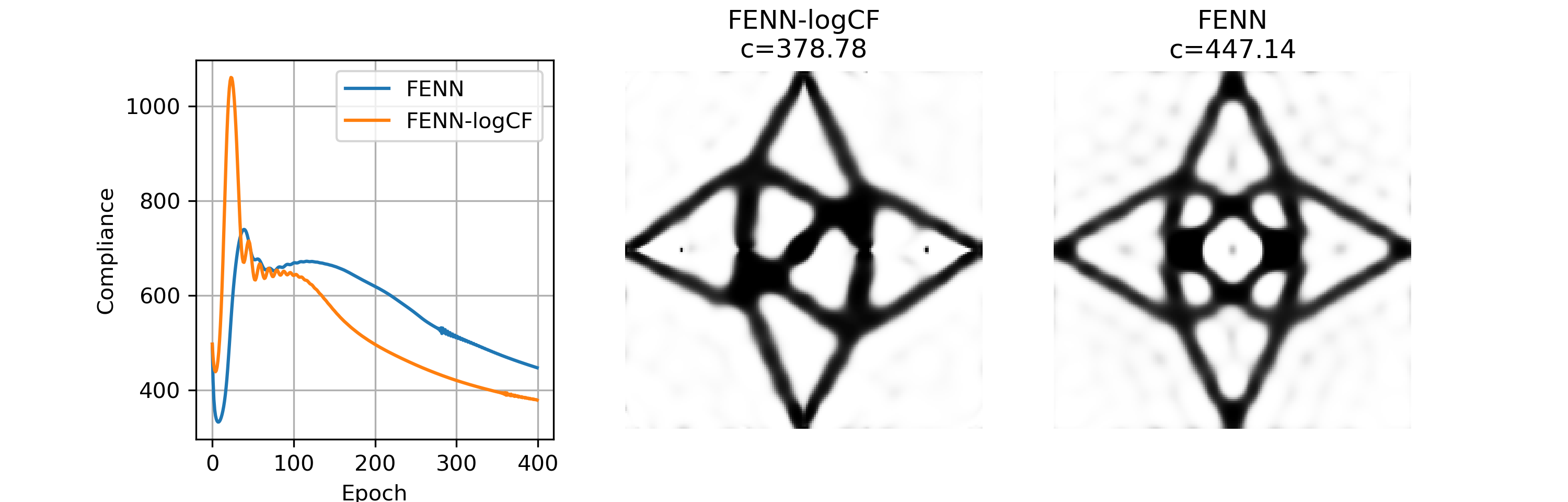}
\end{subfigure}

\caption{We run the same boundary condition for Case 1 with two and three times the resolution. The runtime for 120$\times$120 is 3 min and for $180\times180$ is 20 min}

\label{fig:case1_ss}
\end{figure*}

%% file: fig_3D_CB_03.tex
\begin{figure*}

\centering

\begin{subfigure}[t]{0.25\textwidth}
\includegraphics[width=\textwidth]{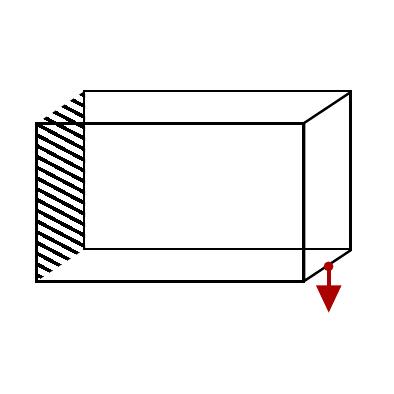}\vspace{-1.0cm}
\caption{Boundary conditions}
\end{subfigure}\begin{subfigure}[t]{0.25\textwidth}
\includegraphics[width=\textwidth]{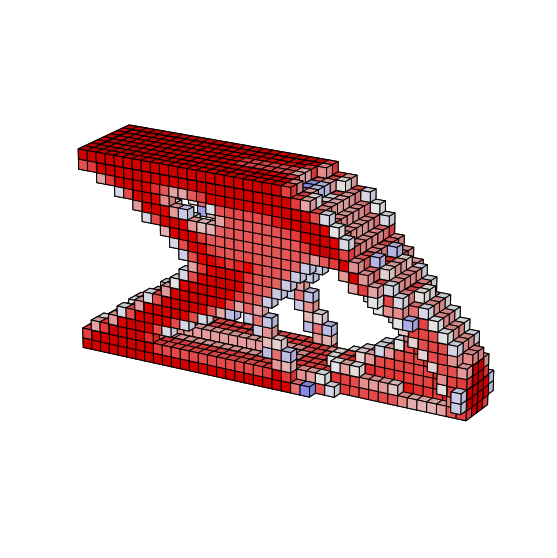}\vspace{-1.0cm}
\caption{Top3d (c = 20.47)}
\end{subfigure}\begin{subfigure}[t]{0.25\textwidth}
\includegraphics[width=\textwidth]{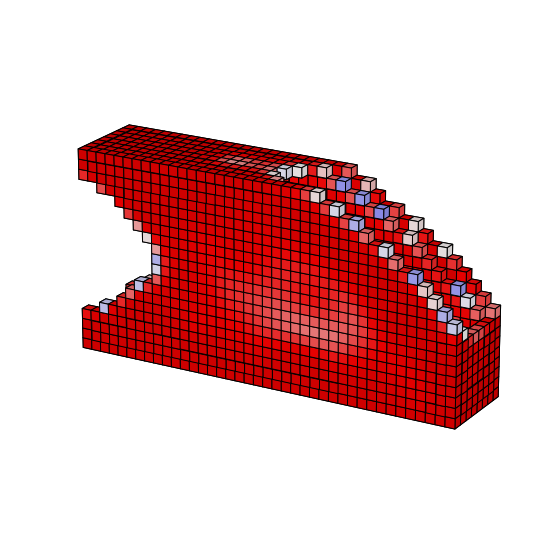}\vspace{-1.0cm}
\caption{FENN (c = 17.44)}
\end{subfigure}\begin{subfigure}[t]{0.25\textwidth}
\includegraphics[width=\textwidth]{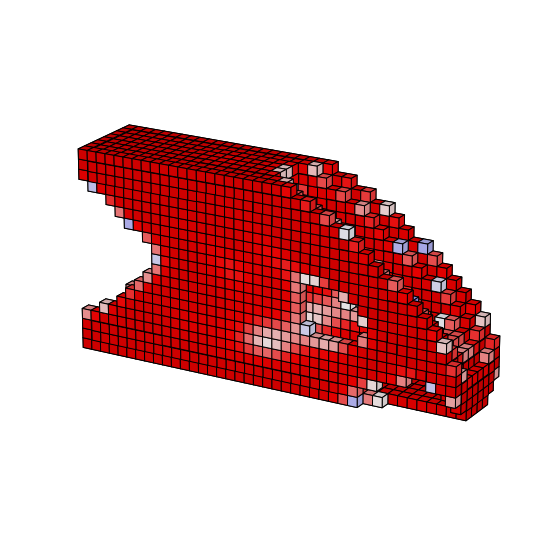}\vspace{-1.0cm}
\caption{FENN-logCF (c = 16.53)}
\end{subfigure}

\vspace{-0.5cm}
\caption{Comparing the results for a 3D cantilever beam example. All examples are run for 200 epochs. b) Top3d \cite{liu2014efficient}(standard 3d topology optimization code using SIMP). c) Using a neural network for density parametrization. d) Using a neural network for density parametrization and additional initial strain energy input with log filtering. We observe that FENN and FENN-logCF choose to create a shell around both side which gives an illusion that the volume fraction is higher. However, the volume fraction is also very close to the target volume fraction of 0.3 (both converged to 0.3003 specifically).}

\label{fig:CB_3D_03}
\end{figure*}

%% file: s5.tex
\section{LIMITATIONS AND FUTURE WORK}
We exploit the ability of neural networks as a universal function approximator to learn the additional mapping from the strain energy conditioning field to the density field output. Currently, the improvement with the conditioning field is not stable across all possible boundary condition configurations. More tuning and testing is required. Another aspect is that the current conditioning field remains fixed during optimization. This is due to the neural network's inability to encode temporal features. The strain energy field changes throughout the optimization, without the ability to capture the temporal feature of the changing strain energy field. As such, the neural network has difficulty providing stable optimization results. 

This work also demonstrates promising results using a conditioning field for online neural topology optimization. The strain energy field may not be the best conditioning field out there and future work may focus on trying out different combinations of conditioning fields similar to TopologyGAN \cite{nie2021topologygan}. This conditioning field approach may demonstrate great synergy with the existing data-driven approach. Using the output of data-driven topology optimization as the conditioning field, online optimization can exploit a conditioning field that is much closer to the final solution. This reduces the complexity of the mapping function for which the neural network needs to learn. Since most data-driven approaches lack the guarantee of compliance minimization, online optimization can serve as the final post-processing step to connect disconnected edges and truly minimize the compliance. 

In this work, we also compare our result against SIMP using "88-lines" \cite{andreassen2011efficient}. However, it may be not possible to determine which one is definitively better or worse. As each program is tuned for different platforms and the possible combinations of problem configuration is endless. Covering all possible problem configurations to reach a conclusion may not be possible. There are exciting possibilities with neural network-based topology optimization, for example, since the design density field is represented by a continuous function, one can infinitely upsample the result to obtain very crisp boundaries \cite{Chandrasekhar2021}. We can also use the same neural network architecture with physics-informed neural networks to conduct mesh-free topology optimization without a FE solver\cite{joglekar2023dmftonn} to name a few. 

%% file: s6.tex
\section{CONCLUSIONS}
We have proposed a novel approach for improving neural network based topology optimization using a conditioning field. Our method involves using a topology neural network that is trained on a case-by-case basis to represent the geometry for a single topology optimization problem. By incorporating the strain energy field calculated on the initial design domain as an additional conditioning field input to the neural network, we have demonstrated faster convergence speed can be achieved. Our results suggest that the efficacy of neural network based topology optimization can be further improved using a prior initial field on the unoptimized domain. We believe that our proposed conditioning field initialization approach could have broad applications in the field of topology optimization, particularly for problems that involve complex geometries.